\pdfoutput=1
\documentclass[lettersize,journal]{IEEEtran}
\usepackage{amsmath, amsfonts}
\usepackage{algorithmic}
\usepackage{algorithm}
\usepackage{array}
\usepackage{textcomp}
\usepackage{stfloats}
\usepackage{url}
\usepackage{verbatim}
\usepackage{graphicx}
\usepackage{cite}
\usepackage{threeparttable}
\usepackage{booktabs}
\usepackage{multirow}
\usepackage{subfigure}
\usepackage{subfig}
\usepackage{booktabs}
\usepackage{threeparttable}
\usepackage{makecell}
\usepackage{multirow}
\usepackage{array}
\usepackage[T1]{fontenc}
\usepackage{etoolbox}
\makeatletter
\patchcmd{\@makecaption}
  {\scshape}
  {}
  {}
  {}
\makeatother
\begin{document}

\title{Heterogeneous Multi-Agent Reinforcement Learning for Zero-Shot Scalable Collaboration}

\author{Xudong Guo, Daming Shi, Junjie Yu, and Wenhui Fan
\thanks{Xudong Guo, Daming Shi, Junjie Yu, and Wenhui Fan are with the Department of Automation, Tsinghua University, Beijing 100084, China (e-mail: gxd20@mails.tsinghua.edu.cn; shidm18@tsinghua.org.cn; yjj23@mails.tsinghua.edu.cn; 
fanwenhui@tsinghua.edu.cn)}
}

\markboth{Heterogeneous MARL for Scalable Collaboration}%
{Shell \MakeLowercase{\textit{et al.}}: A Sample Article Using IEEEtran.cls for IEEE Journals}

\maketitle

\begin{abstract}
The emergence of multi-agent reinforcement learning (MARL) is significantly transforming various fields like autonomous vehicle networks. However, real-world multi-agent systems typically contain multiple roles, and the scale of these systems dynamically fluctuates. Consequently, in order to achieve zero-shot scalable collaboration, it is essential that strategies for different roles can be updated flexibly according to the scales, which is still a challenge for current MARL frameworks. To address this, we propose a novel MARL framework named \textit{S}calable and \textit{H}eterogeneous \textit{P}roximal \textit{P}olicy \textit{O}ptimization \textit{(SHPPO)}, integrating heterogeneity into parameter-shared PPO-based MARL networks. We first leverage a latent network to learn strategy patterns for each agent adaptively. Second, we introduce a heterogeneous layer to be inserted into decision-making networks, whose parameters are specifically generated by the learned latent variables. Our approach is scalable as all the parameters are shared except for the heterogeneous layer, and gains both inter-individual and temporal heterogeneity, allowing SHPPO to adapt effectively to varying scales. SHPPO exhibits superior performance in classic MARL environments like Starcraft Multi-Agent Challenge (SMAC) and Google Research Football (GRF), showcasing enhanced zero-shot scalability, and offering insights into the learned latent variables' impact on team performance by visualization.
\end{abstract}


\def\abstractname{Note to Practitioners}
\begin{abstract}
Multi-agent reinforcement learning (MARL) algorithms are extensively utilized in cooperative multi-agent scenarios, such as multiple rescue unmanned aerial vehicles (UAVs), autonomous vehicles navigating crowded intersections efficiently, and robots jointly managing cargo at ports. However, real-world applications necessitate both scalable and heterogeneous collaboration strategies. For instance, rescue UAV teams may comprise varying numbers of units designated for searching, firefighting, carrying, and communication, depending on the specific tasks. To address this, our approach, SHPPO, can adaptively reorganize roles within the team without requiring additional training when team sizes fluctuate. This superior zero-shot scalability of heterogeneous strategies enables the learned MARL models to be effectively applied to more complex and dynamic tasks. Although a notable limitation of MARL is the disparity between current virtual training environments and real-world scenarios, we are optimistic that advancements in simulation and modeling technologies will soon bridge this gap.
\end{abstract}


\section{Introduction}

\begin{figure}[!t]
\centering
\includegraphics[width=0.5\textwidth]{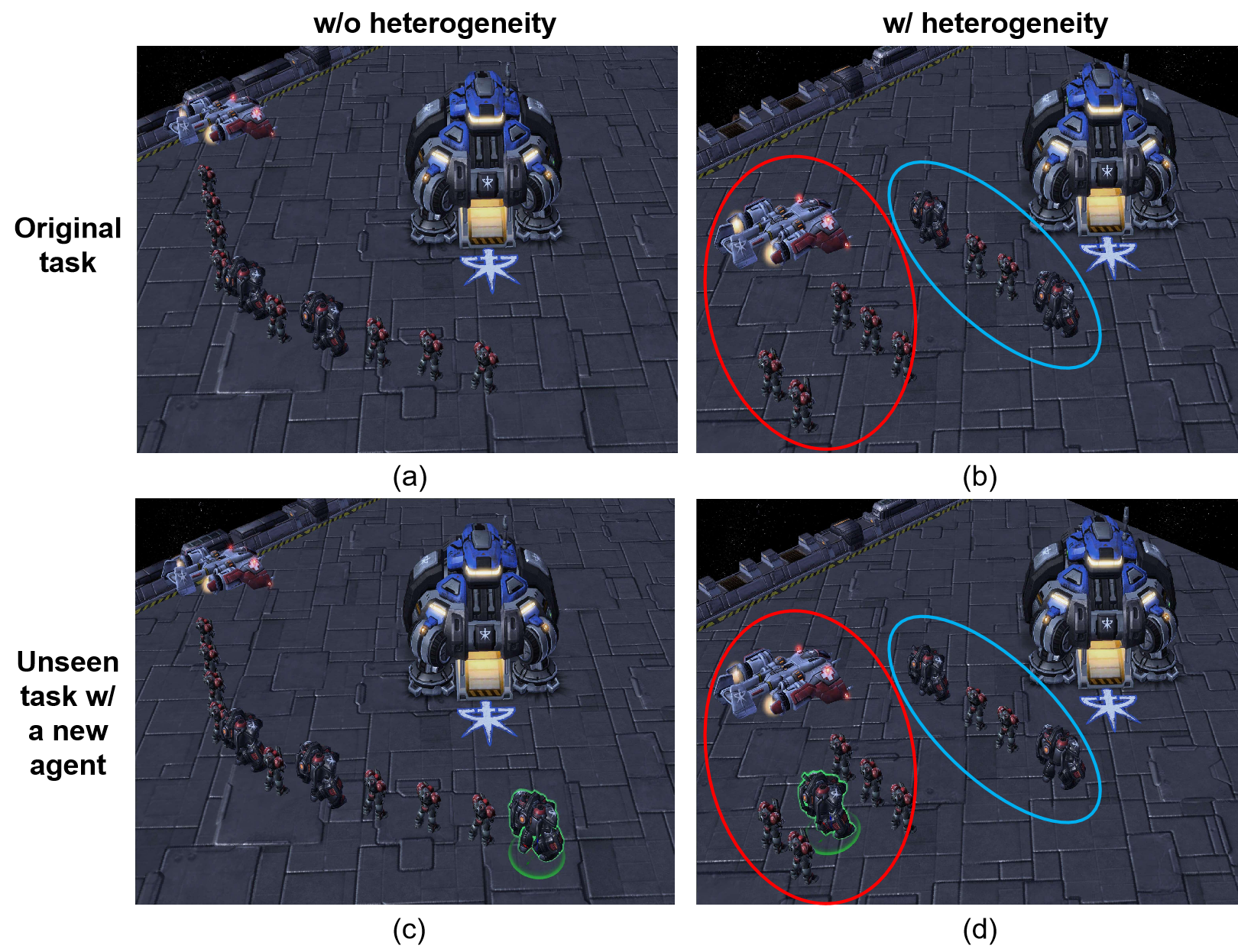}
\caption{\textbf{Illustration of scalability and heterogeneity in SMAC.} (a) The original task without heterogeneous strategies, where different kinds of agents have the same strategy as a circle. (b) The original task with heterogeneous strategies, where the agents form two groups to attract the fire (blue) and attack in the distance (red). (c) The unseen task with a new agent (green) without heterogeneous strategies, where the new agent has the same strategy as others. (d) The new task with heterogeneous strategies, where the new agent can adaptively choose the strategy.}
\label{fig:intro_vis}
\vspace{-0.2in}
\end{figure}

\IEEEPARstart{M}{ulti-agent} systems are playing more and more important roles in future human society with the development of artificial intelligence~\cite{axelrod1981evolution,lesser1998reflections,guo2022pacar,park2023generative,li2023theory}. Multi-agent reinforcement learning (MARL) has seen widespread application in cooperative multi-agent tasks across various domains like autonomous vehicle networks~\cite{sha_languagempc_2023,zhang_cityflow_2019,wang2020multi}, guaranteed display ads~\cite{wang_hierarchical_2021}, and video games~\cite{vinyals_grandmaster_2019}, significantly enhancing collaboration among agents. With the introduction of actor-critic~\cite{konda1999actor} to MARL, recent advances~\cite{lowe2017multi, iqbal2019actor} in MARL exhibit great potential for addressing more complex cooperative tasks.

Despite its successes, traditional MARL methods are confined to specific environments where models are trained. However, in real-world applications, the tasks are usually dynamic and diverse, with issues of \textbf{scalability} (variation of the number of agents) and \textbf{heterogeneity} (diversity of agents' functions and abilities), while it is usually hard to balance the trade-off of scalability and heterogeneity. Simpler mechanisms tend to generalize more effectively across different scales; however, this often comes at the expense of the specificity required for heterogeneous strategies. In such cases, vanilla MARL methods may fail to transfer the learned strategy for a heterogeneous multi-agent system to a new scenario with agents added or removed, as the role assignment and the collaboration mechanism will change. Thus, both scalability and heterogeneity should be considered to design a more practical MARL algorithm.

%

Specifically, in this paper, we focus on zero-shot scalability. This involves directly applying trained MARL models to other unseen tasks with similar settings but varying numbers of agents without the need for further training, which is necessary for various real-world applications. For instance, traffic flows fluctuate throughout the day, requiring a traffic junction coordination algorithm capable of transferring to scenarios with varying numbers of vehicles.


%
Moreover, learning and transferring heterogeneous roles in a team stands out as an important problem of zero-shot scalable collaboration. First, lack of heterogeneity hinders agents from effectively collaborating with diverse counterparts, constraining team performance. As shown in  Fig.~\ref{fig:intro_vis}(a), when heterogeneity is not considered, the agents will have similar strategies and simply form a circle to attack in the game screenshot, while there are different kinds of units with different capacities in the team. But in Fig.~\ref{fig:intro_vis}(b), after the introduction of heterogeneity, some agents may attack in the front to attract the fire (blue circle) while others stay back to be covered. Second, in unseen tasks with a varied number of agents, the role assignment changes with team sizes (refer to our experiments in Fig.~\ref{fig:vis_scal}). An adaptive policy is needed to flexibly assign and adjust agents' heterogeneous roles for better zero-shot scalability. We can implement such a policy based on heterogeneous strategies (Fig.~\ref{fig:intro_vis}(d)), but may fail without heterogeneity (Fig.~\ref{fig:intro_vis}(c)). Therefore, it is significant to answer the question: \textbf{\textit{"How to design a novel MARL framework learning adaptive heterogeneous strategies to achieve better zero-shot scalability?"}}, while the literature only attempted to address the two issues independently.

Some works~\cite{kuba2021trust, li2021celebrating} explore giving each agent a distinct model to improve heterogeneity, but they struggle to scale with the number of agents. They suffer from the dimension curse, and there are no training processes for the newly added agents. While other works~\cite{hu2021updet,zhou2021cooperative} try to build population-invariant MARL for scalability but ignore the heterogeneity. In addition, parameter sharing is scalable and widely used as network parameters for all agents are shared,  bringing benefits of training efficiency and superior performance~\cite{christianos2021scaling,terry2020revisiting,yu2022surprising}, which, however, also limit the diversity and temporal flexibility of the policies. Motivated by these challenges, we aim to enhance zero-shot scalable collaboration by integrating both
 \textbf{inter-individual} (the agents' strategies differ from each other) and \textbf{temporal} (the agents' strategies update with the progress) \textbf{heterogeneity}.

In this paper, we propose a novel MARL framework called \textbf{S}calable and \textbf{H}eterogeneous \textbf{P}roximal \textbf{P}olicy \textbf{O}ptimization \textbf{(SHPPO)} to add a heterogeneous layer to a parameter-shared version of PPO~\cite{2017Proximal} under multi-agent settings (see Fig.~\ref{fig:schematics}). We introduce a latent network to adaptively learn low-dimension latent variables to represent the strategy pattern for each agent, according to its observations and trajectories. Then, based on the latent variable, SHPPO generates the parameters of the heterogeneous layer in the actor network. We also introduce a centralized inference net to guide the learning of the latent network to form a symmetrical structure as the actor-critic in PPO (see Fig.~\ref{fig:schematics}(b)). Note that our approach can be applied to any parameter-shared MARL backbone, and we take PPO as an example to evaluate the performance in this paper.

In this way, though all the network parameters except for the heterogeneous layers are shared for scalability, the heterogeneous layer for every agent is specifically designed to enhance the diversity of the team. In addition to the \textbf{inter-individual heterogeneity}, the agents can update the strategy with the change of observations during the task, so the agents also gain \textbf{temporal heterogeneity}. When transferred to a new scenario with new agents added or removed, the latent variables are accordingly updated with the learned labor division knowledge (See Fig.~\ref{fig:vis_scal}). Thus the following generation of new heterogeneous layers can transfer the learned strategy patterns to achieve zero-shot scalability while keeping heterogeneity.

By conducting extensive experiments on two classic and complex MARL environments Starcraft Multi-Agent Challenge (SMAC)~\cite{samvelyan2019starcraft} and Google Research Football (GRF)~\cite{kurach2020google}, we demonstrate the superior performance of our method over several baselines such as MAPPO~\cite{yu2022surprising} and HAPPO~\cite{kuba2021trust}. Thanks to the heterogeneity, our approach shows better zero-shot scalability when directly transferred to scenarios with varied agent populations. Furthermore, we illustrate the learned latent space with the task progress to analyze how heterogeneity improves the team's performance.


We list the main contributions as follows.
\begin{enumerate}
\item \textbf{New network design}: we introduce actor-critic-like latent net and inference net along with corresponding losses to learn latent representations of the agents' strategy patterns, facilitating the parameter generation of the heterogeneous layer.
    \item \textbf{Novel MARL framework}: the proposed approach can add both inter-individual and temporal heterogeneity to any parameter-shared MARL architecture. The flexible and adaptive heterogeneous strategies help the agents scale to unseen scenarios. 
    
    \item \textbf{Superior performance}: our method SHPPO outperforms the baselines, such as HAPPO and MAPPO, on the original and zero-shot scalable collaboration tasks of SMAC and GRF.
\end{enumerate}


\section{Related Works}
\subsection{MARL for Collaboration}
Multi-Agent Reinforcement Learning (MARL) has emerged as a powerful paradigm for training groups of AI agents to collaborate and solve complex tasks~\cite{sha_languagempc_2023,zhang_cityflow_2019,wang2020multi,wang_hierarchical_2021,vinyals_grandmaster_2019, li2024multi, canzini2024decision, brittain2022scalable}. QMIX~\cite{rashid2020monotonic} addresses the challenge of credit assignment by computing joint Q-values from individual agents as a value-based method, while MADDPG~\cite{lowe2017multi} extends the policy gradient method DDPG~\cite{lillicrap2015continuous} to multi-agent settings using centralized critics and decentralized actors. Different from the off-policy MADDPG, MAPPO~\cite{yu2022surprising}, based on on-policy PPO~\cite{2017Proximal}, achieves strong performance in various MARL benchmarks. Despite significant progress, challenges in MARL persist, including generalizability for unseen scenarios, sample efficiency, non-stationarity, and agent communication~\cite{oroojlooy2023review,du2023review,nguyen2020deep,ijcai2023p15}. In this paper, we mainly focus on zero-shot scalability as one of the generalizability issues.

\subsection{Scaling MARL}
The first step to generalize MARL is to scale the learned MARL models to different population sizes without further training, which is particularly challenging when each agent's policy is independently modeled~\cite{kuba2021trust, ding2020learning}. Some works tackle this by designing population-invariant~\cite{long2020evolutionary}, input-length-invariant~\cite{wang2020few}, and permutation-invariant~\cite{hao2022breaking} networks. Graph neural networks~\cite{agarwal2019learning} and transformers~\cite{zhou2021cooperative, wen2022multi} are adopted to handle the varying populations. UPDeT~\cite{hu2021updet} further decouples the observation and action space, using an importance weight determined with the aid of the self-attention mechanism. In addition, curriculum learning~\cite{long2020evolutionary}, multi-task transfer learning~\cite{qin2022multi}, and parameter sharing~\cite{terry2020revisiting} are also applied to scale MARL. SePS~\cite{christianos2021scaling} models all the policies as $K$ parameter-shared networks, and learns a deterministic function to map the agent to one of the policies, which cannot be adaptively updated during the task. In sum, however, these methods all ignore the heterogeneity or limit the heterogeneity to predefined and fixed $K$ policies when scaling MARL, which may lead to ineffective labor division of the agent teams.

\subsection{Heterogeneous MARL}
 To implement heterogeneity without separately modeling each agent like HAPPO~\cite{kuba2021trust}, HetGPPO~\cite{bettini2023heterogeneous} and CDS~\cite{li2021celebrating}, existing works have delved into this much to learn the role assignment in MARL~\cite{nguyen2022learning, hu2022policy, wang2024romat, hu2024attentionguided}. ROMA~\cite{wang2020roma} tailors individual policies based on roles and exclusively depends on the present observation to formulate the role embedding, which may not fully capture intricate agent behaviors. On the other hand, RODE~\cite{wang2020rode} links each role with a specific subset of the complete action space to streamline learning complexity. LDSA~\cite{yang2022ldsa} introduces heterogeneity from the subtask perspective, but it is hard to determine the total number of subtasks by prior knowledge. Despite these efforts, the existing methods encounter difficulties when scaled to new scenarios with varied population sizes as the number of roles is hard to pre-determine to fit new scenarios, and taking the IDs of agents as inputs to learn the roles makes the integration of a new agent unfeasible. Therefore, achieving a balanced trade-off between scalability and heterogeneity remains an ongoing challenge in MARL.

\section{Background} 
\subsection{DEC-POMDP}
We begin by providing an overview of the background related to Markov Decision Processes (MDP) in multi-agent systems. A Markov game \cite{1994Markov} is defined by a tuple $<N, S, A, P, r, \gamma>$, where $N = \{1, \ldots, n\}$ represents the set of agents, $S$ denotes the finite state space, $A = \prod_{i=1}^{n} A_i$ is the product of finite action spaces for all agents, forming the joint action space. The transition probability function $P: S \times A \times S \rightarrow [0, 1]$ and the reward function $r: S \times A \rightarrow \mathbb{R}$ describe the dynamics of the environment, while $\gamma \in [0, 1)$ is the discount factor. The agents engage with the environment according to the following protocol: at time step $t \in \mathbb{N}$, the agents find themselves in state $s \in S$; each agent $i$ selects an action $a_{i} \in A_i$, drawn from its policy $\pi_i(\cdot | s)$; these individual actions collectively form a joint action $\boldsymbol{a} = (a_{1}, \ldots, a_{n}) \in A$, sampled from the joint policy $\pi(\cdot | s) = \prod_{i=1}^{n} \pi_i(\cdot_i | s)$; the agents receive a joint reward $r = r(s, \boldsymbol{a}) \in \mathbb{R}$ and transition to a new state $s'$ with probability $P(s' | s, \boldsymbol{a})$. The joint policy $\pi$, transition probability function $P$, and initial state distribution $\rho_0$ collectively determine the marginal state distribution at time $t$, denoted by $\rho_t^\pi$.




However, in naive MARL, scalability is hampered by the curse of dimensionality, as the state-action space grows exponentially with the number of agents. Besides, there is the non-stationary problem when agents individually update their policies. To address these two issues, the Centralized Training with Decentralized Execution (CTDE) paradigm \cite{lowe2017multi} is proposed. CTDE allows agents to utilize central information during training, yet during execution each agent acts independently.

When agents are unable to observe the concise state of the environment, commonly referred to as partial observation (PO), a Partially Observable Markov Decision Process (POMDP) extends the MDP model by incorporating observations and their conditional probability of occurrence based on the state of the environment \cite{1998Planning, 2012Reinforcement}. Within the POMDP and CTDE paradigm, the decentralized partially observable Markov decision processes (DEC-POMDP) \cite{2016A} have become a widely adopted model in MARL. A DEC-POMDP is defined by $<N, S, A, O, r, P, \gamma>$.
$o_i = O(s;i)$ represents the local observation for agent $i$ at the global state $s$.
Each agent utilizes its policy $\pi_{\theta_i}(a_i | o_i)$, parameterized by $\theta_i$, to produce an action $a_i$ from the local observation $o_i$. And agents jointly aim to optimize the discounted accumulated reward $J(\theta) = \mathbb{E}_{\boldsymbol{a}, s}[\sum_{t} \gamma^t r(s, \boldsymbol{a})]$, where $\boldsymbol{a}$ is the joint action.


\subsection{Policy Gradient (PG)}
Policy Gradient (PG) reinforcement learning offers the distinct advantage of explicitly learning a policy network, setting it apart from value-based reinforcement learning methods. PG methods aim to optimize the policy parameter $\theta$ to maximize the objective function $J(\theta) = E_S[V_{\pi_\theta}(s)]$.

However, choosing an appropriate learning rate in reinforcement learning proves challenging due to the variance of environments and tasks. Suboptimal learning rates may lead to either slow optimization or value collapse, where updated parameters rapidly traverse the current region in policy space.

To address this challenge and ensure stable policy optimization, Trust Region Policy Optimization (TRPO) \cite{2015Trust} introduces a constraint on the parameter difference between policy updates. This constraint restricts parameter changes to a small range, preventing value collapse and facilitating monotonic policy learning. 

Denote the advantage function as $A(s,\boldsymbol{a}) = Q(s,\boldsymbol{a}) - R$, with $R = \sum_{t} \gamma^t r(s, \boldsymbol{a})$. In the training episode $k$, the parameter update in TRPO follows $\mathop\theta_{k+1}={\arg\max}_{\theta} L(\theta_k,\theta)~s.t. \bar{D}_{KL}(\theta||\theta_k)\leq\delta$, where $L(\theta_k,\theta) = E[\frac{\pi_\theta(a|s)}{\pi_{\theta_k}(a|s)}A_{\pi_{\theta_k}}(s,a)]$ approximates the original policy gradient objective $J(\theta)$ within the constraint of KL divergence.

Building upon TRPO, Proximal Policy Optimization (PPO) \cite{2017Proximal} offers a simplified version that maintains the learning step constraint while being more computationally efficient and easier to implement.

In PPO, the objective function is given by:
\begin{equation}
\begin{aligned}
\label{eq:PPO_actor}
\mathcal{L}(s,a,&\theta_k,\theta)=\mathbb{E}
\left[\min\left(\frac {\pi_\theta(a|s)}{\pi_{\theta_k}(a|s)} A_{\pi_{\theta_k}}(s,a), \right.\right.\\
&\left.\left.\text{clip}\left(\frac {\pi_\theta(a|s)}{\pi_{\theta_k}(a|s)},1-\epsilon,1+\epsilon\right)A_{\pi_{\theta_k}}(s,a)\right)\right],
\end{aligned}
\end{equation}
forcing the ratio $\frac {\pi_\theta(a|s)}{\pi_{\theta_k}(a|s)}$ to be within the interval $(1-\epsilon,1+\epsilon)$. This ensures that the new parameter $\theta$ remains close to the old parameter $\theta_k$.

\subsection{PG in Multi-agent}
Multi-agent PPO (MAPPO) extends PPO to the multi-agent scenario~\cite{yu2022surprising}, specifically considering DEC-POMDP.
MAPPO employs parameter sharing among homogeneous agents, where agents share the same network structure and parameters during both training and testing. MAPPO is a CTDE framework, with each PPO agent maintaining a parameter-shared actor network $\pi_\theta$ for policy learning and a critic network $V(s,\phi)$ for value learning, where $\theta$ and $\phi$ are the parameters of the policy and value networks, respectively. The value function requires the global state and is used during training to reduce variance.

MAPPO faces challenges when dealing with heterogeneous agents or tasks and lacks the monotonic improvement guarantee present in trust region learning. To address these limitations, \cite{kuba2021trust} introduced Heterogeneous-Agent TRPO (HATRPO) and Heterogeneous-Agent PPO (HAPPO) algorithms. These algorithms leverage the multi-agent advantage decomposition lemma and a sequential policy update scheme to ensure monotonic improvement. HATRPO/HAPPO extend single-agent TRPO/PPO methods to the context of multi-agent reinforcement learning with heterogeneous agents.

Building on the multi-agent advantage decomposition lemma, HAPPO calculates the actor loss for each agent as follows:
\begin{equation}
\begin{aligned}
\label{eq:HAPPO_actor}
\mathcal{L}_{Actor}(\theta)=\mathbb{E}
\left[\min \left( \frac {\pi_{\theta^{i_m}}^{i_m}(a^i|s)}{\pi _{\theta _k^{i_m}}^{i_m}(a^i|s)} M^{i_{1:m}}(s,\boldsymbol{a}), \right. \right.\\
\left. \left.\text{clip}\left( \frac {\pi_{\theta^{i_m}}^{i_m}(a^i|s)}{\pi _{\theta _k^{i_m}}^{i_m}(a^i|s)}, 1 \pm \epsilon \right) M^{i_{1:m}}(s,\boldsymbol{a}) \right) \right],
\end{aligned}
\end{equation}
where $i_{1:m}$ denotes the permutation of updated $m$ agents, $\theta _k^{i_m}$ is the network parameter of agent $i_m$ in training episode $k$, and $M^{i_{1:m}}(s,\boldsymbol{a})=\prod_{i=1}^{i_{1:m}} \frac {\pi_{\theta^{i}}^{i}(a^i|s)}{\pi _{\theta _k^{i}}^{i}(a^i|s)}(V(s,\phi)- R)$ represents the weighted advantage in HAPPO.

Both MAPPO and HAPPO calculate the critic loss as follows:
\begin{equation}
\begin{aligned}
\label{eq:HAPPO_critic}
\mathcal{L}_{Critic}(\phi)=\mathbb{E}
\left[(V(s,\phi)- R)^2 \right].
\end{aligned}
\end{equation}

In our work, we adopt the parameter-shared HAPPO as the baseline and backbone for constructing our approach, SHPPO.

\section{Scalable and Heterogeneous \\ Multi-Agent Reinforcement Learning}

\begin{figure*}[t]
\centering
\includegraphics[width=\textwidth]{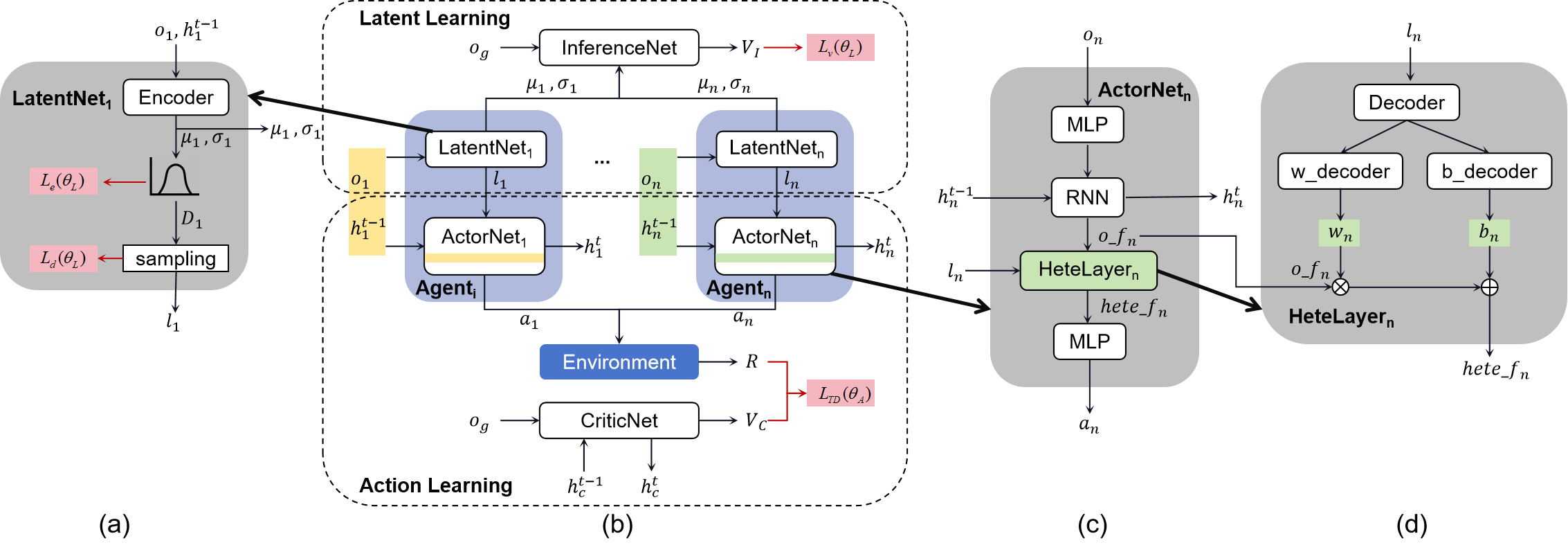}
\caption{\textbf{Schematics of our approach SHPPO.} (a) Architecture of LatentNet to represent the agent's strategy pattern as a low-dimensional latent variable $l$. (b) Framework of SHPPO to conduct latent learning and action learning with dual actor-critic networks. Each agent contains two networks ActorNet and LatentNet. The two networks are parameter-shared among the agents except for the heterogeneous layer (HeteLayer), so SHPPO is scalable. (c) Architecture of ActorNet with HeteLayer, bringing heterogeneity in SHPPO. (d) Architecture of HeteLayer where the latent variable $l$ is decoded to the parameters of the heterogeneous linear layer. The pink blocks denote the losses. The yellow and green bars denote different HeteLayers.}
\label{fig:schematics}
\vspace{-0.1in}
\end{figure*}

Existing heterogeneous MARL methods, such as HAPPO, model each agent as an independent and different network; it suffers from its inflexibility and huge parameter costs when the scale of the multi-agent system increases. Especially, when new agents are introduced, there is no pre-trained model available for them, as each model is unique. In this section, we propose a novel MARL framework SHPPO, which can learn adaptive heterogeneous strategies that can be transferred for zero-shot scalable collaboration.

We try to solve this problem in another direction different from HAPPO: we add adaptive heterogeneity to one single parameter-shared model. Specifically, as in Fig. \ref{fig:schematics}(b), we structure the new framework into two components: latent learning and action learning. First, we implement a latent learning model (LatentNet) to adaptively generate a latent distribution $\mathcal{D}$ for each agent based on its observation and memory (in practice, we take the hidden state of the RNN in the actor network (ActorNet) as the memory). The learned latent distribution represents the strategy pattern of the specific agent, which is further sampled and applied to the computation of the heterogeneous layer parameters for this agent. Except for the heterogeneous layer, the other layers' parameters in the ActorNet are shared by all the agents. When the number of agents varies, the LatentNet can always generate a unique latent distribution for each agent so that the proposed method can handle the scalability and heterogeneity at the same time. Our approach is agnostic to the implementation of the action learning part, and in practice, we adopt the shared version of HAPPO as the backbone for its monotonic improvement property.

In the following subsections, we first introduce the latent learning (Section~\ref{sec:latent}) and heterogeneous layers (Section~\ref{sec:hete}). Then, we elaborate on the loss design (Section~\ref{sec:loss}) and the efficacy of the mechanism (Section~\ref{sec:efficacy}). Finally, we conclude with the overall training procedure (Section~\ref{sec:overall}).

\subsection{Scalable Latent Learning}
\label{sec:latent}
To design a heterogeneous layer for each agent, we first need to represent the agent's strategy pattern as a low-dimensional latent variable $l$ as shown in Fig. \ref{fig:schematics}(a). Take agent $i$ as an example, we take a 3-layer MLP as the Encoder to convert the observation $o_i$ and the memory from the RNN $h^{t-1}_i$ at time step $t$ to a latent distribution. Specifically, the Encoder outputs the mean $\boldsymbol{\mu}_i$ and the standard deviation $\boldsymbol{\sigma}_i$ of the multi-dimensional Gaussian $\boldsymbol{\mathcal{D}}_i$. Like humans, the agents' strategy patterns may not be exactly the same under similar situations, so the Encoder learns a distribution instead of a fixed vector. The final latent variable $l_i$ is generated by sampling to keep the randomness as 
\begin{equation}
\label{eq:latent}
l_i \sim \boldsymbol{\mathcal{N}}(\boldsymbol{\mu}_i, \boldsymbol{\sigma}_i).
\end{equation}

Inspired by the classic Actor-Critic architecture~\cite{konda1999actor}, an inference net (InferenceNet) is introduced as a critic to guide the learning of the LatentNet (see Fig.~\ref{fig:schematics}(b)). Though the LatentNet generates the latent variable individually for each agent, the InferenceNet centrally learns an overall value function to evaluate the whole team. Specifically, the InferenceNet takes  $\boldsymbol{\mu}$ and $\boldsymbol{\sigma}$ from all the agents as the inputs, together with the joint global observation $o_g$, and then computes $V_{I}$ as the evaluation for the LatentNet. The InferenceNet is trained by supervised learning to minimize the difference between $V_{I}$ and the return from the environment $R$. In another way, it judges whether the learned latent distributions are good enough to help with decision-making based on global observations. Thus, the LatentNet's goal is to maximize the $V_{I}$.

It is noteworthy that all the agents share the same parameters. Thanks to the CTDE architecture, during the execution and transfer, we only forward the LatentNet and ActorNet, namely, the Agent part in gray color in Fig.~\ref{fig:schematics}(b). Though the input dimensions of the InferenceNet and CriticNet vary with the population sizes, this will not influence the transfer to the new scenario to execute. In this way, our method is scalable to fit scenarios with different agent numbers. Please refer to Appendix Table~\ref{tab:net} for the detailed network settings.

\subsection{Heterogeneous Layer Design}
\label{sec:hete}
With the generated latent variable for each agent, we can further design the heterogeneous layer for each agent described in Fig.~\ref{fig:schematics}(c, d). The heterogeneous layer is a linear layer in the ActorNet. The ActorNet for agent $i$ takes the observation $o_i$ and the hidden states of the RNN $h^{t-1}_i$ to get the action $a_i$. Here, in order to make better sequential decisions, the RNN is introduced to keep the memory from the previous steps by the hidden states. The ActorNet for different agents shares the parameters, except for the heterogeneous layer, whose parameters are generated with each agent's $l$. 
The heterogeneous layer computes $hete\_f_i$ as Equation \ref{eq:hete_f}. 

\begin{equation}
\label{eq:hete_f}
hete\_f_i = w_i * o\_f_i + b_i.
\end{equation}

Here, the weights $w_i$ and the bias $b_i$ are distinct for each agent, resulting in heterogeneous decision-making policies. A decoder, implemented as an MLP, along with two separate linear decoders (denoted as w\_decoder and b\_decoder), generates $w_i$ and $b_i$ based on $l_i$. In sum, the LatentNet learns to represent the strategy patterns, and the heterogeneous layer turns the latent variable into part of the ActorNet's parameters. This allows agents to exhibit unique action styles, despite sharing the majority of parameters. Furthermore, they can adapt their styles in response to evolving tasks or changes in team size. The execution of one time step is shown in Algorithm~\ref{alg:alg1}.

\begin{algorithm}[htb]
\caption{Execution of one time step $t$.}\label{alg:alg1}
\begin{algorithmic}
\STATE 
\textbf{Input: }observations $\boldsymbol{o}$ and hidden states from the last time step $\boldsymbol{h}^{t-1}$ for all the agents, LatentNet and ActorNet sharing parameters among the agents.
\STATE
\STATE{\textbf{Distributed execution}

\STATE
\hspace{0.5cm} Compute latent variable $l_i$ by LatentNet with the inputs of $o_i, h_i^{t-1}$ for each agent.
\STATE \hspace{0.5cm} Compute the parameters $w_i, b_i$ for the HeteLayer in each agent's ActorNet with the input $l_i$.
\STATE \hspace{0.5cm} Generate the action $a_i$ for each agent by ActorNet with $o_i, h_i^{t-1}$ as inputs.
\STATE \hspace{0.5cm} Store the hidden state $h_i^t$ for the next step.
}
\STATE
\STATE{\textbf{Interaction with the environment}

\STATE \hspace{0.5cm} Agents execute the the actions $\boldsymbol{a}$ in the environment together.

\STATE \hspace{0.5cm} The environment returns the reward $r$ and the flag whether this episode is finished.
}

\end{algorithmic}
\label{alg1}
\end{algorithm}

\subsection{Loss Formulation}
\label{sec:loss}
In this subsection, we formulate the losses to train the networks deployed in our framework.
A good latent representation should be helpful for decision-making, and 
 also identifiable and diverse. Therefore, we design three distinct loss items to guide the learning of the LatentNet.

We first employ the value $V_I$ from the InferenceNet $\mathcal{L}_v$ to evaluate the impacts of the latent variables on decision-making:

\begin{align}
\label{eq:L_v}
\mathcal{L}_v(\theta_{L}) = &V_I(o_g, \boldsymbol{\mu}(\boldsymbol{o}, \boldsymbol{h}^{t-1}; \theta_{L}), \boldsymbol{\sigma}(\boldsymbol{o}, \boldsymbol{h}^{t-1}; \theta_{L})),
\end{align}



where $\boldsymbol{\mu}$ and $\boldsymbol{\sigma}$ are the concatenation of each agent's $\boldsymbol{\mu}_i$ and $\boldsymbol{\sigma}_i$. Note that when calculating $\mathcal{L}_v(\theta_{L})$ to update the LatentNet, the parameters of the InferenceNet are fixed and will not be updated. To maximize $\mathcal{L}_v$, we can expect that the learned latent variables will be improved to help the agents choose a better heterogeneous layer.

Then, we incorporate two additional unsupervised losses as regularization terms to facilitate the learning of identifiable and diverse latent variables. 

The first item is $\mathcal{L}_e$, the mean entropy of the $\boldsymbol{\mathcal{D}}_i$,  defined as follows:
\begin{align}
\label{eq:L_e}
\mathcal{L}_e(\theta_{L}) &= \frac{1}{n}\sum_{i=1}^{n}\mathcal{H}(\mathcal{D}_i(o_i, h^{t-1}_i; \theta_{L})) \notag \\
&= \frac{1}{n}\sum_{i=1}^{n}\mathcal{H}(\mathcal{N}(\boldsymbol{\mu}_i(o_i, h^{t-1}_i; \theta_{L}), \boldsymbol{\sigma}_i(o_i, h^{t-1}_i; \theta_{L})))
,
\end{align}
where $n$ is the team size, $\mathcal{H}$ is the entropy function of the distribution, $\theta_{L}$ is the parameters of the LatentNet. To minimize the $\mathcal{L}_e(\theta_{L})$, the latent distributions tend to be sharper and thus more identifiable.

The second item is the distance among the agents' latent variables $\mathcal{L}_d(\theta_{L})$:
\begin{align}
\label{eq:L_d}
\mathcal{L}_d(\theta_{L}) = &\frac{1}{n(n-1)}\sum_{i=1}^{n}\sum_{j\neq i}^{n}Norm(1 - \notag \\
&cos\_sim(l_i(o_i, h^{t-1}_i; \theta_{L}), l_j(o_j, h^{t-1}_j; \theta_{L}))),
\end{align}

where $cos\_sim$ is the cosine similarity between two vectors. $Norm$ is the normalization of all the distances to ensure the distance distribution :
\begin{equation}
\label{eq:norm}
Norm(x_i)|_{x_i \in \boldsymbol{x}} = \frac{x_i-min(\boldsymbol{x})}{max(\boldsymbol{x})-min(\boldsymbol{x})+10^{-12}}.
\end{equation}
By maximizing the $\mathcal{L}_d(\theta_{L})$, the latent variables will be diverse to encourage the agents to have heterogeneous styles. Note we implement the sampling function as rsample() in Pytorch to get reparameterized samples, so the sampling is differentiable.

Up to this point, we have the overall loss for the LatentNet $\mathcal{L}_L(\theta_{L})$ to be minimized as follows,
\begin{equation}
\label{eq:L_L}
\mathcal{L}_L(\theta_{L}) = -\mathcal{L}_v(\theta_{L}) + \lambda_e\mathcal{L}_e(\theta_{L}) - \lambda_d\mathcal{L}_d(\theta_{L}),
\end{equation}
where $\lambda_e$ and $\lambda_d$ are the weights of the regularization items. Please see Appendix Table~\ref{tab:para} for the details of the weights.


The InferenceNet is learned by a mean squared error (MSE) loss between the value $V_I$ and environment return $R$:

\begin{align}
\label{eq:L_I}
\mathcal{L}_I(\theta_{I}) = &MSE\_Loss[V_I(o_g, \boldsymbol{\mu},  \boldsymbol{\sigma}; \theta_{I}), R],
\end{align}


where the MSE loss is calculated as 
\begin{equation}
\label{eq:MSE}
MSE\_Loss[X, Y] = \frac {1}{B}\sum_{k=1}^{B}(x_{k} - y_{k})^{2},
\end{equation}
where $B$ is the minibatch size, $x_{k}$ and $y_{k}$ are the samples in the minibatch.
The InferenceNet will be able to infer how good the latent distributions are when $\mathcal{L}_I$ converges. Then, The value $V_I$ will be a good estimation of the return.

Besides, the ActorNet and the CriticNet are updated following the HAPPO loss Equation~\ref{eq:HAPPO_actor} and Equation~\ref{eq:HAPPO_critic}, including the parameters of the decoder, w\_decoder, and b\_decoder.
The loss of the ActorNet is a Temporal Difference (TD) loss,
$\mathcal{L}_{TD}$:
\begin{align}
\label{eq:L_TD}
\mathcal{L}_{TD}(\theta_{A}) &=  \mathcal{L}_{Actor}(s,\boldsymbol{a},\theta_{A})\notag \\
&= \mathcal{L}_{Actor}(V_C, R, \boldsymbol{a}(\boldsymbol{o}, \boldsymbol{h}^{t-1}, \boldsymbol{l}; \theta_{A})),
\end{align}
note that here $\boldsymbol{l}$ is taken as an input without gradients. $V_C$ is the value from the critic.

The loss for the critic is 
$\mathcal{L}_{C}$:
\begin{align}
\label{eq:L_C}
\mathcal{L}_{C}(\theta_{C}) &=  \mathcal{L}_{Critic}(s, \theta_{C})\notag \\
&= \mathcal{L}_{Critic}(V_{C}(o_g, h^{t-1}_{c}; \theta_{C}), R).
\end{align}


\subsection{Efficacy of the LatentNet and InferenceNet}
\label{sec:efficacy}
In this subsection, we further explain how the latent learning works to adaptively represent the agents' heterogeneous strategies in a low-dimension space as the latent variables. 
\subsubsection{InferenceNet}

The InferenceNet is trained to approximate a value function \( V_I \) based on the latent representations of all the agents $\boldsymbol{\mu}(\boldsymbol{o}, \boldsymbol{h}^{t-1}; \theta_{L})$ and $ \boldsymbol{\sigma}(\boldsymbol{o}, \boldsymbol{h}^{t-1}; \theta_{L})$ produced by the LatentNet, as well as the global observation \( o_g \). The objective is to minimize the difference between the predicted value \( V_I \) and the true return \( R \). This difference is captured by the MSE loss function (Equation~\ref{eq:L_I}).

The gradient of this loss function $\mathcal{L}_{I}$ with respect to the parameters \( \theta_I \) of the InferenceNet is:
\begin{align}
\label{eq:L_I_2}
\nabla_{\theta_I} \mathcal{L}_{I}(\theta_{I}) &= 2\mathbb{E}[(V_I - R) \nabla_{\theta_I} V_I].
\end{align}

By iteratively applying gradient descent, the parameters \( \theta_I \) are adjusted such that the value function \( V_I \) better approximates the true return \( R \). Namely, the value \( V_I \) centrally assesses the overall adequacy of the learned latent distributions of the entire team in facilitating decision-making processes based on global observations.



\subsubsection{LatentNet}
With the value \( V_I \) as one of the optimization targets, the LatentNet learns better latent representations to maximize the value as Equation~\ref{eq:L_L}.
The gradient of this loss function $\mathcal{L}_{v}$ with respect to the parameters \( \theta_L \) of the LatentNet is given by:
\begin{align}
\label{eq:L_L2}
\nabla_{\theta_L} \mathcal{L}_{v}(\theta_{L}) &= \nabla_{\theta_L} V_I(o_g, \mu(o, h_{t-1}; \theta_L), \sigma(o, h_{t-1}; \theta_L))\notag \\
&= \frac{\partial V_I}{\partial \mu} \cdot \frac{\partial \mu}{\partial \theta_L} + \frac{\partial V_I}{\partial \sigma} \cdot \frac{\partial \sigma}{\partial \theta_L}.
\end{align}

Here, \( \frac{\partial V_I}{\partial \mu} \) and \( \frac{\partial V_I}{\partial \sigma} \) represent the sensitivity of the value function \( V_I \) to the changes in the latent representations \( \mu \) and \( \sigma \), respectively. Meanwhile, \( \frac{\partial \mu}{\partial \theta_L} \) and \( \frac{\partial \sigma}{\partial \theta_L} \) denote the dependence of these latent representations on the parameters of the LatentNet. By maximizing this loss, the LatentNet learns to produce latent vectors \( \mu \) and \( \sigma \) that are more effective in representing the policy, thereby enhancing the overall decision-making process.



In conclusion, the output of the InferenceNet, which serves as the loss for the LatentNet, establishes a direct feedback mechanism that aligns the latent representations with the objective of maximizing the expected value function, resulting in enhanced policy performance. Additionally, when combined with the two other unsupervised losses, $\mathcal{L}_e$ and $\mathcal{L}_d$, the latent representations become more identifiable and diverse.

\begin{figure*}[!t] 
    \centering
    \includegraphics[width=\textwidth]{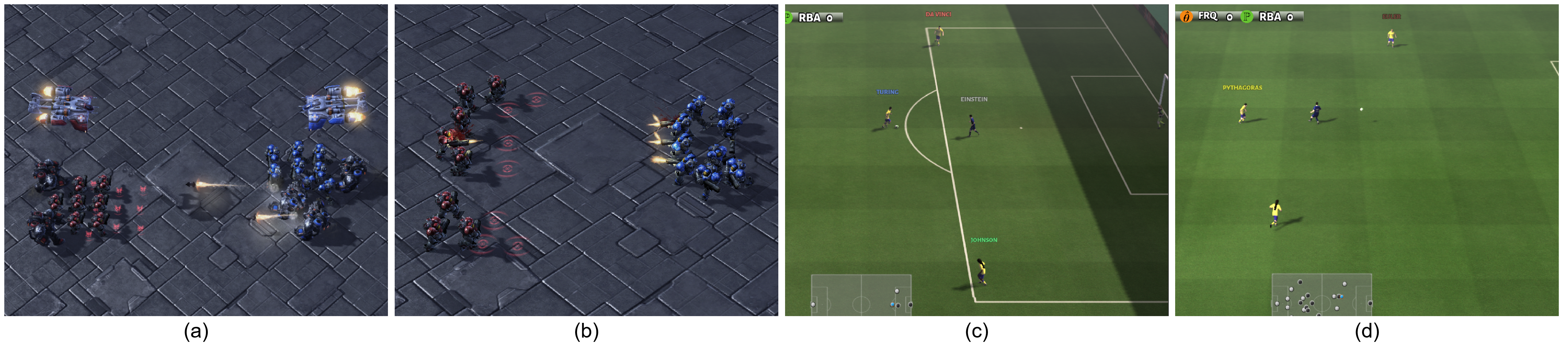} 
  \caption{\textbf{Visualization of the tasks.} (a) MMM2 in SMAC. (b) 8m\_vs\_9m in SMAC. (c) 3\_vs\_1\_with\_keeper in GRF. (d) counterattack \_easy in GRF.}
  \label{fig:task_vis}
\end{figure*}

\subsection{Overall Training Procedure}
\label{sec:overall}
In one episode, the first phase is the interaction with the environment to collect data. Every agent gains a latent variable and then decides its action decision at each step (see Algorithm~\ref{alg:alg1}). Data such as observations, hidden states, latent variables, actions, and rewards are saved in the buffer. After the agents succeed, fail, or the step limit is reached, the second phase is training (see Algorithm~\ref{alg:alg2}). The agents update their ActorNets one by one, following the procedure in HAPPO \cite{kuba2021trust}. Of course, here all the agents share the same ActorNet. Thus, the ActorNet will be updated for $n$ times based on different agents' observations, actions, and rewards. After that, the CriticNet, the LatentNet, and the InferenceNet update their parameters. Please refer to Appendix Table~\ref{tab:net} and \ref{tab:para} for the detailed network and hyperparameter configurations.

\begin{algorithm}[ht]
\caption{Overall Training Procedure.}\label{alg:alg2}
\begin{algorithmic}
\STATE 
\textbf{Input:} batch size $B$, number of agents $n$, number of episodes $K$, max steps per episode $T$.
\STATE 
\textbf{Initialize:} parameter-shared networks: ActorNet $\{\theta_A\}$, CriticNet $\{\theta_C\}$, LatentNet $\{\theta_L\}$,
InferenceNet $\{\theta_I\}$, replay buffer $\mathcal{B}$.
\STATE{\textbf{for $k=0:K$ do}

\STATE
\hspace{0.5cm} Collect trajectories and latent distributions.
\STATE \hspace{0.5cm} Push transitions $\{(o^t_i, h^{t-1}_i, l^t_i, a^t_i, o^{t+1}_i, r^t)\}, 0 \leq t \leq T, i=1, ..., n$ into $\mathcal{B}$.
\STATE \hspace{0.5cm} Sample a random minibatch of $B$ transitions from $\mathcal{B}$.
\STATE \hspace{0.5cm} Draw a random permutation of agents $i_{1:n}$.
\STATE\hspace{0.5cm} {\textbf{for agent $i_m=i_1, ..., i_n$ do}
\STATE \hspace{1cm} Update ActorNet as Equation~\ref{eq:L_TD} with $o_{i_m}, h^{t-1}_{i_m}, l_{i_m}$
}
\STATE\hspace{0.5cm} \textbf{end for}
\STATE\hspace{0.5cm} Update CriticNet as Equation~\ref{eq:L_C}.

\STATE\hspace{0.5cm} Compute 
 $\mathcal{L}_v(\theta_{L})$ by InferenceNet as Equation~\ref{eq:L_v}.
 
\STATE\hspace{0.5cm} Compute $\mathcal{L}_e(\theta_{L}), \mathcal{L}_d(\theta_{L})$ by forwarding LatentNet with all the agents' $\boldsymbol{o}$ and $\boldsymbol{h^{t-1}}$ as Equation~\ref{eq:L_e} and Equation~\ref{eq:L_d}.

\STATE\hspace{0.5cm} Update LatentNet by Equation~\ref{eq:L_L}: \STATE\hspace{1cm} $\mathcal{L}_L(\theta_{L}) = -\mathcal{L}_v(\theta_{L}) + \lambda_e\mathcal{L}_e(\theta_{L}) - \lambda_d\mathcal{L}_d(\theta_{L})$.
\STATE\hspace{0.5cm} Update InferenceNet by $\mathcal{L}_I(\theta_{I})$ as Equation~\ref{eq:L_I}.
\STATE\textbf{end for}
}
\end{algorithmic}
\label{alg1}
\end{algorithm}

\section{Experiments}
\subsection{Environments and Metrics}

To demonstrate the effectiveness of our method, especially when scaled to unseen scenarios, we adopt two classic and hard MARL environments - Starcraft Multi-Agent
Challenge (SMAC)~\cite{samvelyan2019starcraft} and Google Research Football (GRF)~\cite{kurach2020google}, and further modify them to design scalability tasks. 

StarCraft II is a real-time strategy game serving as a classic benchmark in the MARL community. In the Starcraft Multi-Agent
Challenge (SMAC) tasks, a team of allied units aims to defeat the adversarial team in various tasks. 

Google Research Football (GRF) contains diverse subtasks in a football game. The agents cooperate as part of a football team to score a goal against the opposing team under scenarios such as counterattack, shooting in the penalty area, etc.

\begin{figure*}[!t] 
    \centering
    \includegraphics[width=\textwidth]{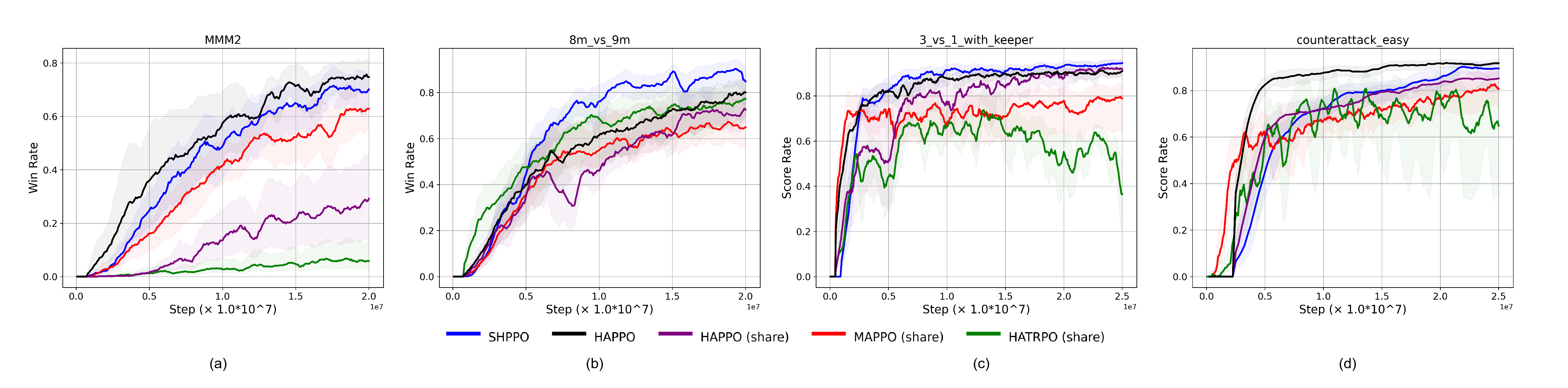} 
    \caption{\textbf{Performance on SMAC and GRF.} We plot the win rate on SMAC and the score rate on GRF during training. (a) MMM2 (SMAC). (b)  8m\textunderscore vs\textunderscore9m (SMAC). (c) 3\textunderscore vs\textunderscore1\textunderscore with\textunderscore keeper (GRF). (d) counterattack\textunderscore easy (GRF). The confidence interval is calculated over 5 seeds.} 
    \label{fig:performance} 
\end{figure*}

\begin{table*}[!ht]
\caption{\textbf{Win/score rate and reward on SMAC and GRF.}
}
\label{tab:performance}
\resizebox{\linewidth}{!}{
	\centering\tiny
	\begin{threeparttable}
		\begin{tabular}{c c r@{$\pm$}l r@{$\pm$}l r@{$\pm$}l r@{$\pm$}l r@{$\pm$}l}
			\toprule
			 Task & Metric & \multicolumn{2}{c}{SHPPO (ours)}& \multicolumn{2}{c}{HAPPO} & \multicolumn{2}{c}{HAPPO (share)*}& \multicolumn{2}{c}{MAPPO (share)}&\multicolumn{2}{c}{HATRPO (share)} \\
			\midrule
\multirow{2}{*}{MMM2} & Win rate & 71.2& 6.5 & \textbf{76.3}& \textbf{5.1} &31.2&20.4 &62.7&10.1 & 6.8& 5.9   \\
& Reward & 17.5& 1.3 & \textbf{17.8}& \textbf{1.1} &14.7& 3.1 & 16.9& 1.5 & 8.9& 1.2  \\
\midrule
\multirow{2}{*}{8m\_vs\_9m} & Win rate & \textbf{85.5}& \textbf{5.8} & 81.0&10.5 & 70.5& 9.3 &65.2&12.8& 78.0& 8.8\\
& Reward & \textbf{18.9}&\textbf{0.8} & 18.3& 1.2 & 17.2& 1.6 & 17.5& 1.9 & 17.5& 1.5 \\

   \midrule\midrule
			\multirow{2}{*}{3\textunderscore vs\textunderscore1\textunderscore with\textunderscore keeper}  & Score rate & \textbf{94.2}& \textbf{2.1} & 90.8& 2.4 & 91.3&3.7 & 78.7&11.5 &41.3&20.8   \\
   
   & Reward & \textbf{19.9}& \textbf{0.1} & 19.6& 0.2 & 19.7& 0.3 & 17.0& 1.1 & 13.3& 1.3 \\
			 \midrule
			\multirow{2}{*}{counterattack\textunderscore easy }  & Score rate &91.2& 3.0 & \textbf{92.0}& \textbf{4.1} & 86.4& 5.8 & 81.8& 7.3 & 61.4&22.5\\
   
    & Reward & 18.9& 0.2 & \textbf{19.0}& \textbf{0.3} & 18.5& 0.4 & 17.3& 0.5 & 16.5& 1.3 \\

\bottomrule
		\end{tabular}
  \begin{tablenotes} 
            \item[*] Implemented based on HAPPO but every agent shares the same parameters. 
        \end{tablenotes}
	\end{threeparttable}
}

\end{table*}

In order to directly zero-shot transfer the learned models to tasks with varied populations, we modify the original environments to fix the number of allies and enemies that one agent can observe. This way, the observations and actions length is the same before and after the transfer. Hence, we can concentrate on transferring strategies without the need to map observations, which is not the primary focus of this paper. More details of the modified tasks can be found in Appendix~\ref{appendix:env}.
All the experiments in our paper are conducted on the new scalable tasks.
	
As for metrics, we adopt training rewards to evaluate the training progress. We also test the performance on the test environments to calculate the win rate for SMAC or score rate for GRF.

We compare the performance of our approach with four other baselines. As SHPPO is built based on parameter-shared HAPPO, we take both original HAPPO where each agent has an individual network, and HAPPO (share) where every agent shares the same parameters as baselines. HATRPO~\cite{kuba2021trust} is another heterogeneous MARL method, and for a fair comparison, we adopt its parameter-shared version as a baseline. MAPPO~\cite{yu2022surprising} is a classic homogeneous MARL method, and in this paper, we also use parameter-shared MAPPO.

\begin{figure*}[t]    
  \centering  \includegraphics[width=0.75\textwidth]{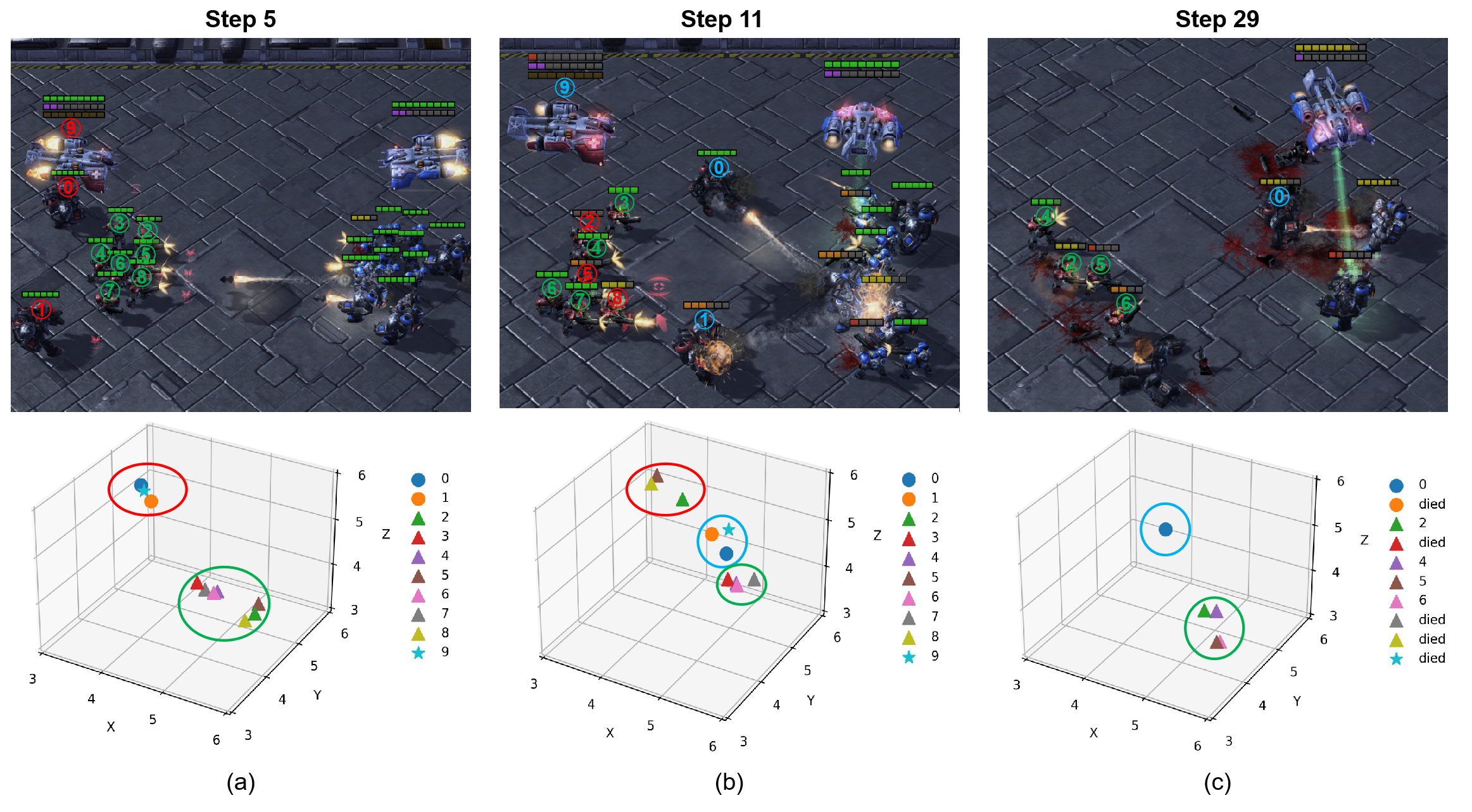}
  \caption{\textbf{Visualization of MMM2 and latent variables.} The agents adaptively employ heterogeneous strategies throughout the task. The latent variables are clustered based on these strategies, indicated by circles of different colors: red: stay back to be covered. green: attack in the distance. blue: attract the fire by moving to the front. Each agent is identified by an ID, and their types are distinguished by the shapes in the lower figures: solid circles: Marauders, triangle: Marines, star: Medivac.}
  \label{fig:vis_latent}  
\end{figure*}

\begin{table*}[!h]
\caption{\textbf{Win rate of scalability test on SMAC.} 
}
\label{tab:scal_smac}
\resizebox{\linewidth}{!}{
	\centering\tiny
	\begin{threeparttable}
		\begin{tabular}{c r@{$\pm$}l r@{$\pm$}l r@{$\pm$}l r@{$\pm$}l r@{$\pm$}l}
			\toprule
			 Task* & \multicolumn{2}{c}{SHPPO (ours)}& \multicolumn{2}{c}{HAPPO**} & \multicolumn{2}{c}{HAPPO (share)}& \multicolumn{2}{c}{MAPPO (share)}&\multicolumn{2}{c}{HATRPO (share)} \\
			\midrule
   \textbf{MMM2 (721\_831)***} & 71.2&6.5 & \textbf{76.3}&\textbf{5.1} &31.2&20.4 & 62.7&10.1 &6.8&5.9   \\
              
              711\textunderscore731 & \textbf{5.6}&\textbf{2.5}& 2.5&1.8 & 0.0& 0.0 & 5.1& 2.5& 0.0& 0.0  \\
              621\textunderscore821 & \textbf{77.5}&\textbf{7.8}& 56.8& 11.2 & 25.1& 12.5 & 75.2& 12.5& 9.5& 5.6  \\
              621\textunderscore631 & \textbf{62.5}&\textbf{2.5}& 35.1& 22.4 & 26.4& 14.1 & 62.5& 15.9& 2.6& 2.1  \\
              722\textunderscore831 & \textbf{96.3}&\textbf{1.5}& 87.5& 10.1 & 95.6& 2.5 & 96.1& 3.8& 45.4& 5.2  \\

    821\textunderscore831 & 82.5& 7.5& 12.5& 4.9 & 70.0& 19.2 & \textbf{95.1}& \textbf{4.5}& 18.7& 7.5  \\
              731\textunderscore831 & 95.6& 4.5& 31.8& 30.2 & 96.4& 3.5 & \textbf{97.5}&\textbf{2.5}& 41.6& 7.7  \\
			\midrule\midrule
   \textbf{8m\textunderscore vs\textunderscore9m} & \textbf{85.5}&\textbf{5.8} & 81.0&10.5 & 70.5&9.3 &65.2& 12.8&78.0  &8.8\\
            6m\textunderscore vs\textunderscore7m & \textbf{17.5}& \textbf{10.4}& 2.5& 1.2 & 12.7& 9.8 & 15.7& 11.9& 12.6&8.2   \\
            7m\textunderscore vs\textunderscore8m & \textbf{42.5}& \textbf{10.7}& 2.5&2.5 & 40.5& 10.2 & 37.1& 23.5& 37.5& 12.1  \\
            10m\textunderscore vs\textunderscore11m & \textbf{70.2}&\textbf{20.1}& 0.0& 0.0 & 51.3& 23.8 & 42.5& 18.3 & 68.1& 22.5\\
\bottomrule
		\end{tabular}
  \begin{tablenotes} 
            \item[*] The first row is the original task, while the others are unseen tasks with varied numbers of agents, whose results are achieved by zero-shot transfer. 
            \item[**] When transferring HAPPO, we randomly select a model of the same type for the added agent, or remove extra models when there are fewer units in the new task.
            \item[***] We use the format ABC\_DEF to represent the team sizes of both sides. A, B, C are the number of Marines, Marauders, Medivacs of the ally side controlled by our algorithm. D, E, F are the number of Marines, Marauders, Medivacs of the enemy side. The original MMM2 can be represented as 721\_831.
        \end{tablenotes}
	\end{threeparttable}
}
\end{table*}

\subsection{Results on SMAC}

We first evaluate SHPPO on the original tasks. The first task is a heterogeneous task MMM2, where each side has three kinds of different units - Marine, Marauder, and Medivac (Fig.~\ref{fig:task_vis}(a)). They have different health points, hit points, and even special skills. Therefore, heterogeneous strategies will improve the team's performance by maximizing each agent's advantages. In Fig.~\ref{fig:performance}(a), HAPPO performs best as it models every agent individually. However, thus HAPPO consumes more parameters and lacks scalability. In contrast, our approach, SHPPO, outperforms all the other parameter-shared baselines without modeling each agent. The training reward and win rate of SHPPO are very close to those of HAPPO, showing that SHPPO has a similar heterogeneous representation ability to HAPPO (See Table~\ref{tab:performance}). The performance of the two heterogeneous methods, HAPPO and HATRPO, is the worst when parameters are shared.

We also illustrate the screenshots with the corresponding latent variables (Fig.~\ref{fig:vis_latent}). Surprisingly, the locations of the latent variables can explain the agents' strategies. At the early stage of the task (Step 5), the agents form two groups, the Marines move forward and attack in the front (green circle) to cover the Marauders and the Medivac behind (red circle). At step 11, the two armies encounter and the strategies alter accordingly. The Marauders get close to the enemies to attract the fire while the Medivac keeps curing them (blue circle). Some Marines (Agent 2, 5, and 8) are hurt with low health points, thus they retreat to be covered (red circle), while the other Marines attack the enemies (green circle). In the final stage, some agents are dead and the enemies are almost defeated. The only Marauder keeps attracting the fire in the front. However, the Marines attack the enemies regardless of their health points to achieve final victory. We can see that the latent variables successfully learn the heterogeneous strategy patterns in this task, which are located diversely in the latent space. These latent variables contribute to the adaptive policy updates during the task through the HeteLayer in the ActorNet, so the agents can not only have distinct strategies but also change their strategies as the task progresses.

Moreover, we evaluate the proposed method on a homogeneous task 8m\_vs\_9m (Fig.~\ref{fig:task_vis}(b)), where each side has the same kind of unit - Marine. Though the agents are all the same, they can have different strategies and adaptively alter during the task. As shown in Fig.~\ref{fig:performance}(b) and Table~\ref{tab:performance}, SHPPO outperforms all the baselines including HAPPO. The training process of SHPPO also converges faster than other baselines.

\subsection{Results on GRF}

We test the performance on two tasks of GRF. In task 3\_vs\_1\_with\_keeper (Fig.~\ref{fig:task_vis}(c)), our algorithm can control three football players against one adversary player and the keeper. SHPPO is the most effective method among all the methods in Fig.~\ref{fig:performance}(c) and Table~\ref{tab:performance}.

In another task counterattack\_easy (Fig.~\ref{fig:task_vis}(d)), four agents need to cooperate together to do a counterattack. They need to pass the ball against one adversary player and the keeper to score. In Fig.~\ref{fig:performance}(d), SHPPO reaches a similar performance with HAPPO after training, though SHPPO does not explicitly model each agent. Still, SHPPO outperforms all the parameter-shared baselines (also see Table~\ref{tab:performance}).

\begin{figure*}[t]    
  \centering  
  \includegraphics[width=0.75\textwidth]{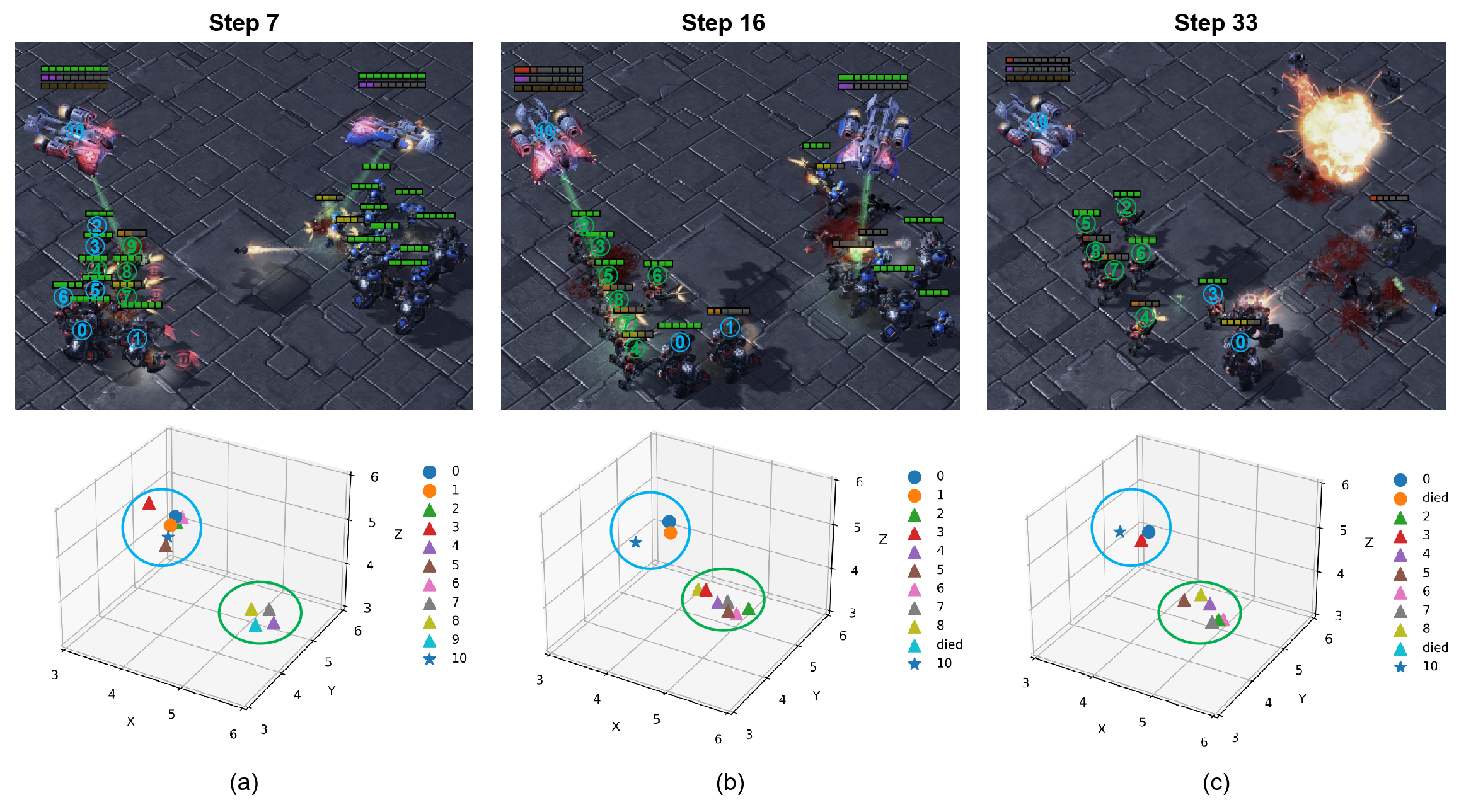}
  \caption{\textbf{Visualization of the transferred task 821\_831 and latent variables.} The heterogeneous strategies update accordingly when the population size and task difficulty vary. The notations are consistent with those in Fig.~\ref{fig:vis_latent}.}
  \label{fig:vis_scal}  
\end{figure*}

\begin{table*}[!htbp]
\caption{\textbf{Score rate of scalability test on GRF.}
}
\label{tab:scal_grf}
\resizebox{\linewidth}{!}{
	\centering\tiny
	\begin{threeparttable}
		\begin{tabular}{c r@{$\pm$}l r@{$\pm$}l r@{$\pm$}l r@{$\pm$}l r@{$\pm$}l}
			\toprule
			 Task & \multicolumn{2}{c}{SHPPO (ours)}& \multicolumn{2}{c}{HAPPO} & \multicolumn{2}{c}{HAPPO (share)}& \multicolumn{2}{c}{MAPPO (share)}&\multicolumn{2}{c}{HATRPO (share)} \\
			\midrule
   \textbf{3\textunderscore vs\textunderscore1\textunderscore with\textunderscore keeper}  & \textbf{94.2}&\textbf{2.1} & 90.8&2.4 & 91.3&3.7 & 78.7&11.5 &41.3&20.8   \\
			 4\textunderscore vs\textunderscore1\textunderscore with\textunderscore keeper & \textbf{90.1}& \textbf{9.5}& 82.3& 10.7 & 47.5& 20.6 & 61.2& 15.3& 37.8& 21.5  \\
    2\textunderscore vs\textunderscore1\textunderscore with\textunderscore keeper & \textbf{92.3}&\textbf{7.7}& 91.0& 3.5 & 91.3& 6.4 & 63.4& 22.1& 45.2& 17.3  \\
    
			\midrule\midrule
   \begin{tabular}[c]{@{}c@{}} \textbf{counterattack\textunderscore easy*}\\\textbf{(4\textunderscore vs\textunderscore1\textunderscore counterattack\textunderscore with\textunderscore keeper)}\end{tabular}  & 91.2&3.0&\textbf{92.0}&\textbf{4.1} & 86.4&5.8
  &81.8 &7.3&61.4& 22.5\\
  
			  5\textunderscore vs\textunderscore1\textunderscore counterattack\textunderscore with\textunderscore keeper & 87.8&3.7 & \textbf{89.2}&\textbf{4.9}  &  86.5&3.4  &77.1&5.7 &68.2&12.3\\
    3\textunderscore vs\textunderscore1\textunderscore counterattack\textunderscore with\textunderscore keeper & \textbf{27.4}&\textbf{4.1} & 1.0&1.0  &  24.7&5.2  &15.1&6.5 &23.4&5.7\\

\bottomrule
		\end{tabular}
  \begin{tablenotes} 
            
            \item[*] The original task counterattack\_keeper can be represented as 4\textunderscore vs\textunderscore1\textunderscore counterattack\textunderscore with\textunderscore keeper. The algorithm can control 4 players to counterattack against 1 adversary player and the adversary keeper. 
        \end{tablenotes}
	\end{threeparttable}
}

\end{table*}	

\subsection{Sacability Test}

The proposed method SHPPO not only performs well on the original tasks on SMAC and GRF, but also has extraordinary zero-shot scalability. We construct a series of new tasks with different numbers of agents and enemies to test the scalability.

In the original MMM2 task, the ally team has 7 Marines, 2 Marauders, and 1 Medivac, while the enemy team has 8 Marines, 3 Marauders, and 1 Medivac. This task is labeled as 721\_831. To adjust the difficulty of the new task appropriately, we either add one ally unit or remove one ally unit and one enemy unit together. This ensures that the task is neither too difficult nor too easy, avoiding situations where any method would have a win rate of 0\% or 100\%.

When transferring the learned model from the original task to the unseen tasks, SHPPO does not require additional training. While HAPPO cannot be directly transferred due to each agent having a distinct model, we address this by randomly selecting a model of the same type for the added agent, or removing extra models when there are fewer units in the new task.

From Table~\ref{tab:scal_smac}, SHPPO performs the best on most of the unseen zero-shot tasks. Even in cases where SHPPO is not optimal, it still performs comparably to the best. In tasks 821\_831 and 731\_831, there is one more Marine and Marauder, respectively. Therefore, the homogeneity increases after the transfer, while MAPPO is trained with a shared network and may work better when the task is more homogeneous. It's worth noting that although we attempt to transfer HAPPO, its performance on all the new tasks is worse due to over-fitting to the role assignments in the original task, despite having a better win rate on the original task.

In Fig.~\ref{fig:vis_scal}, we illustrated the latent variables after the transfer (821\_831), where one new Marine is added while the enemies are the same, making the new task easier. Thus, the agents inherit the previous strategies and roles in the team, but further adaptively align them with the easier scenario. At step 7, the added Marine (Agent 9) joins its fellow Marines to attack the enemies (green circle), while other Marines are moving forward to attract and disperse the enemies' fires, together with the Medivac and Marauders (blue circle). Note that no agents keep staying back to be covered as the red circle in Fig.~\ref{fig:vis_latent}, because they are taking a more aggressive strategy in easier tasks. At step 16, all the Marines stop moving and focus on attacks (green circle). The Marauders go on moving to the front with the support of the Medivac (blue circle). At the final stage (step 33), Marauder 1 dies, and almost at the same time, Marine 3 changes its strategy to replace the dead Marauder, moving to the front (blue circle). With the good representation ability of the latent variables, the HeteLayer can inherit the learned strategies in the original task. More importantly, thanks to the adaptive inference of latent variables based on the observations, the agents are able to update collaboration strategies accordingly when the population size and task difficulty vary.

\begin{figure*}[ht] %
    \centering
    \includegraphics[width=\textwidth]{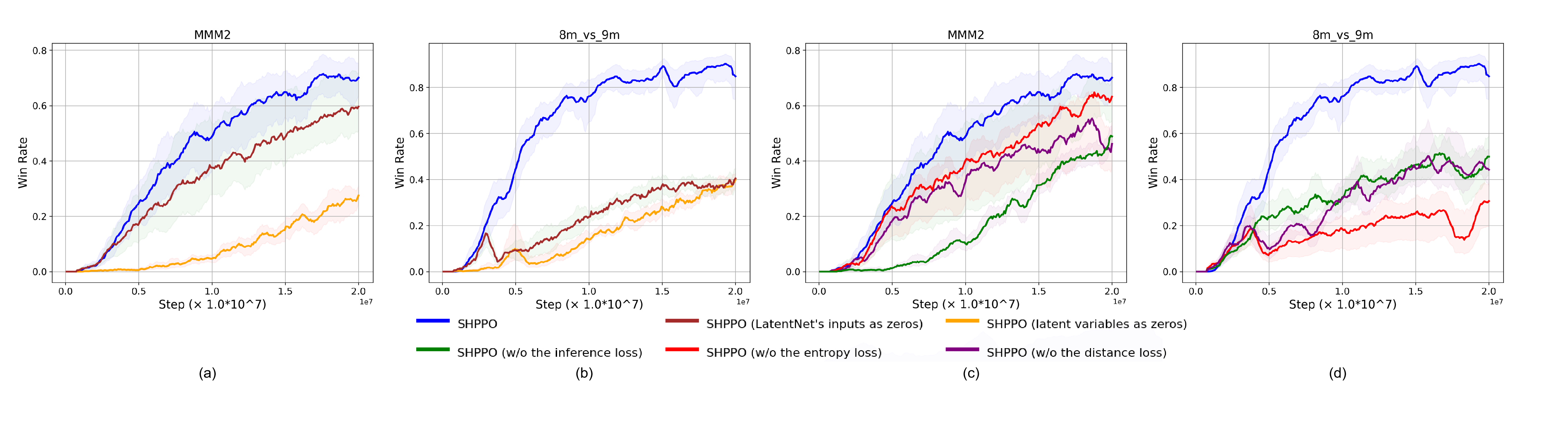} 
    \caption{\textbf{Ablation studies.} The experiments are conducted on MMM2 and 8m\textunderscore vs\textunderscore9m in SMAC. (a, b) Ablation studies of LatentNet's inputs. (c, d) Ablation studies of LatentNet's losses. The confidence interval is calculated over 5 seeds.} 
    \label{fig:ablation} 
\end{figure*}

On another task 8m\_vs\_9m of SMAC, we also create new tasks with both fewer and more agents, such as 6m\_vs\_7m and 10m\_vs\_11m. SHPPO is superior in all the unseen tasks. Surprisingly, HAPPO almost cannot win after the transfer. We suppose that though the agents can have distinct roles in a homogeneous task, their strategies may not be extremely different as they belong to the same unit type. HAPPO possibly overfits every agent's heterogeneous policy to task 8m\_vs\_9m, which may not be suitable for the new tasks. Therefore, when we try to transfer HAPPO, the learned strategies do not work on the unseen tasks.

Similarly, we test scalability by removing or adding one agent in the two tasks of GRF. SHPPO maintains the best performance after transfer on the 3\textunderscore vs\textunderscore1\textunderscore with\textunderscore keeper series. 
For the counterattack\_easy series, despite HAPPO's best performance on the original task, SHPPO achieves the superior score rate on 3\_vs\_1\_counterattack\_with\_keeper after the transfer. When adding a new agent, SHPPO also has comparable performance.

\subsection{Ablation Studies}

We further conduct ablation studies to examine the effects of different parts of latent learning. 

In order to test the effectiveness of the latent variables, we first only mask the inputs of the LatentNet with zeros so that the latent variables cannot be adaptively updated according to the new observations. Moreover, in another experiment, we replace all the latent variables with zeros to totally remove the latent learning. In Fig.~\ref{fig:ablation}(a, b), when the inputs of LatentNet are zeros, the performance on two tasks of SMAC reduces significantly, especially in task 8m\_vs\_9m. When the latent variables are set as zeros, it is equal to the removal of the latent learning part, and we can see even worse performance in such a case. Therefore, flexible adjustment of heterogeneous strategies is important for the agents to cooperate.

Next, we test the effectiveness of the losses of latent learning. We first remove the guidance from the InferenceNet $\mathcal{L}_v(\theta_{L})$ while keeping the other losses $\mathcal{L}_e(\theta_{L})$ and $\mathcal{L}_d(\theta_{L})$. Results in Fig.~\ref{fig:ablation}(c, d) show that without $\mathcal{L}_v(\theta_{L})$, there will be a performance decline on both MMM2 and 8m\_vs\_9m. It suggests that the InferenceNet is able to assess the value of latent variables to help with agents' decision-making. 

Similarly, in Fig.~\ref{fig:ablation}(c, d), we then keep $\mathcal{L}_v(\theta_{L})$ but  remove $\mathcal{L}_e(\theta_{L})$ and $\mathcal{L}_d(\theta_{L})$, respectively. The two experiments have similar performance reductions on task MMM2. Nevertheless, on the homogeneous task 8m\_vs\_9m, the experiment without the entropy loss $\mathcal{L}_e(\theta_{L})$ has the lowest win rate. We believe this is due to the fact that, while strategies for a homogeneous task may exhibit some degree of diversity for labor division, they cannot be excessively random or divergent, as the agents belong to the same unit type. While the entropy loss $\mathcal{L}_e(\theta_{L})$ can make sure that the learned latent variables are relatively deterministic and identifiable.

\section{Conclusion}
In this paper, we propose a novel MARL framework Scalable and Heterogeneous Proximal Policy Optimization (SHPPO) to integrate both inter-individual and temporal heterogeneity to any parameter-shared MARL backbone, so that the framework can learn adaptive heterogeneous strategies to achieve zero-shot scalable collaboration. We introduce latent learning to adaptively update the parameters of the heterogeneous layer and the agents' strategies according to the transferred scales. Our approach shows superior performance on both original tasks and scalability tests over baselines in several classic MARL tasks. Future work could implement the heterogeneous layer with more complex network designs like transformers, or integrate population-invariant methods to further enhance scalability. One can even take advantage of the emerging large language models (LLMs), converting the latent variables to heterogeneous prompts for LLM agents to gain better scalability.


 

{\appendices
\section{Technical Appendix}

\label{appendix:env}
\textbf{Environment settings.} We modify and keep the number of observations the same in one task series, so that we can focus on the transfer of strategies without mapping the observations. For example, in the task MMM2 of SAMC, there are 10 controlled agents and 12 enemies, respectively. In the original tasks of SMAC, the agent can observe all the enemies, resulting in easy and unscalable tasks. We modify the tasks to limit the observation range to the closest 10 enemies and the closest 8 allies so that we can create a series of new tasks with different numbers of agents and enemies to test scalability. Similarly, in  8m\_vs\_9m, we limit the range to the closest 6 enemies and the closest 5 allies. While in GRF, there are always 11 players in one team. We change the number of controlled players to test scalability but keep the observations as the global observations of all the players in the field.





\label{appendix:net}
\textbf{Training details.} We provide the network configurations in Table~\ref{tab:net} and hyperparameters used for training in Table~\ref{tab:para}. For each task, we train SHPPO on one single NVIDIA V100 GPU. We conduct 20 million training steps for the SMAC environments and 25 million training steps for the GRF environments. The training time is shown in Table~\ref{tab:time}.
\begin{table}[!h]
\caption{\textbf{Network configurations for SHPPO.}
}
\label{tab:net}
	\centering
	\begin{threeparttable}
		\begin{tabular}{cc}
			\toprule
			 Module & Implementation \\
			\midrule
   MLP dim & 64 \\
   Encoder & 3-layer MLP \\
   Decoder & 3-layer MLP \\
    InferenceNet & 3-layer MLP \\
   w\_decoder & linear layer \\
   b\_decoder & linear layer \\
   HeteLayer & linear layer \\
  
   ActorNet RNN hidden dim & 64 \\
 CriticNet RNN hidden dim & 64 \\
 activation & ReLU \\
  optimizer & Adam \\
  latent variable dim & 3 \\

\bottomrule
		\end{tabular}
	\end{threeparttable}

\vspace{-0.2in}
\end{table}	

\label{appendix:para}
\begin{table}[h]
\caption{\textbf{Hyperparameters settings for SHPPO.}
}
\label{tab:para}
	\centering
	\begin{threeparttable}
		\begin{tabular}{cc}
			\toprule
			 Parameter & Value \\
			\midrule
   mini batch num & 1 \\
   ActorNet learning rate &0.0005\\
   CriticNet learning rate &0.0005\\
   LatentNet learning rate &0.0005\\
   InferenceNet learning rate & 0.005\\
   entropy loss weight $\lambda_e$ &0.01\\
   distance loss weight $\lambda_d$ &0.1\\
   discount factor $\gamma$ & 0.95 \\
   evaluate interval & 25 \\
   evaluate times & 40 \\
   clip & 0.2 \\
   GAE lambda & 0.95 \\
   max steps per episode & 160\\

\bottomrule
		\end{tabular}
	\end{threeparttable}

\vspace{-0.2in}
\end{table}		

\begin{table}[h]
\caption{\textbf{Training Time of SHPPO.}
}
\label{tab:time}
	\centering
	\begin{threeparttable}
		\begin{tabular}{cc}
			\toprule
			 Task & Time (h) \\
			\midrule
   MMM2 & 25.5$\pm$0.2 \\
   8m\_vs\_9m & 20.3$\pm$0.6\\
   3\_vs\_1\_with\_keeper &26.1$\pm$0.8\\
   counterattack\_easy &15.4$\pm$0.5 \\

\bottomrule
		\end{tabular}
	\end{threeparttable}

\vspace{-0.2in}
\end{table}


 

\bibliography{main}
\bibliographystyle{IEEEtran}

\begin{IEEEbiography}[{\includegraphics[width=1in,height=1.25in,clip,keepaspectratio]{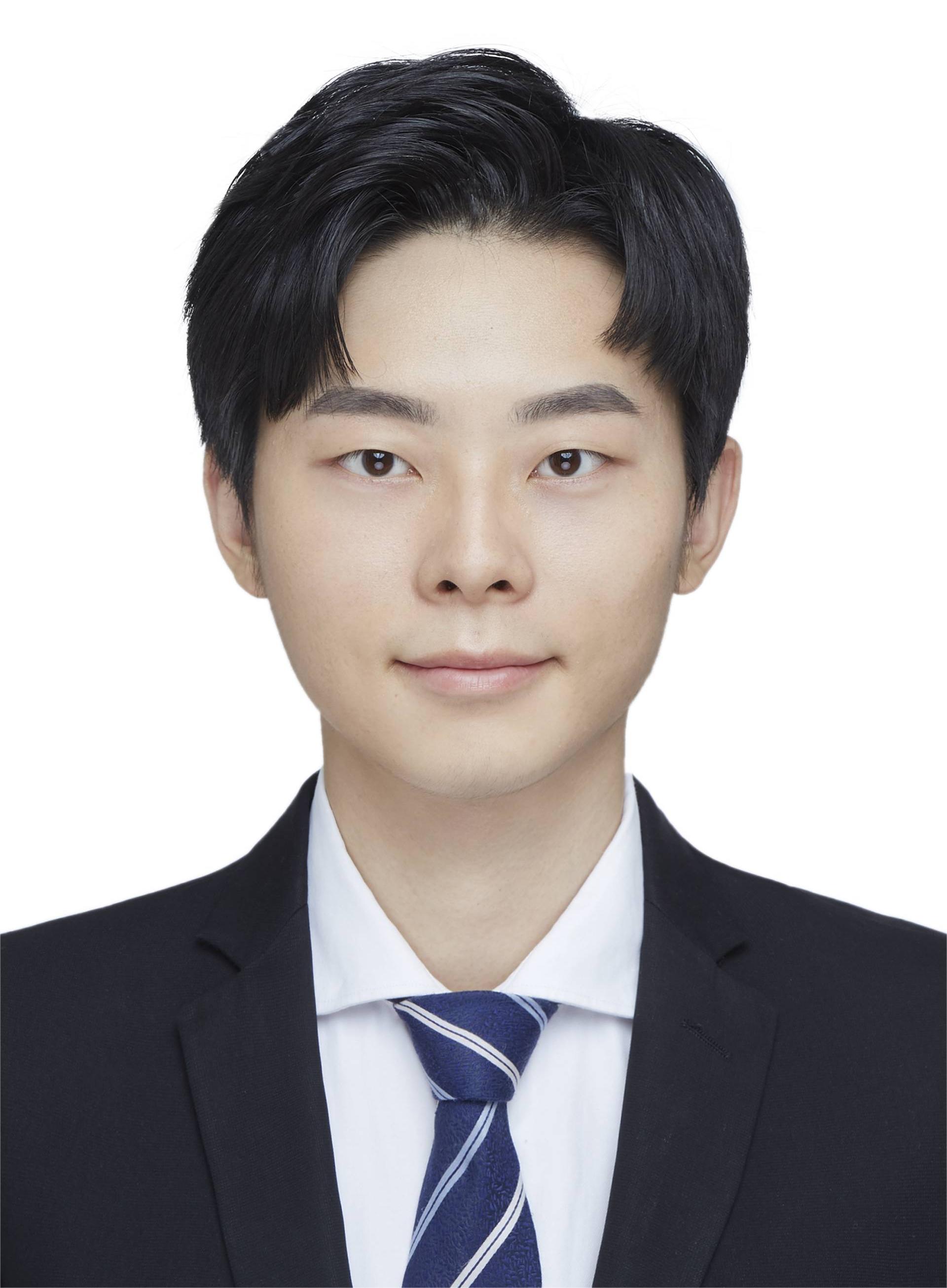}}]{Xudong Guo}
received the B.Eng. degree in automation from Tsinghua University, Beijing, China, in 2020. He is now pursuing the Ph.D. degree at the Department of Automation, Tsinghua University, Beijing, China. His research interests include multi-agent systems and reinforcement learning.
\end{IEEEbiography}

\vspace{-0.2in}
\begin{IEEEbiography}[{\includegraphics[width=1in,height=1.25in,clip,keepaspectratio]{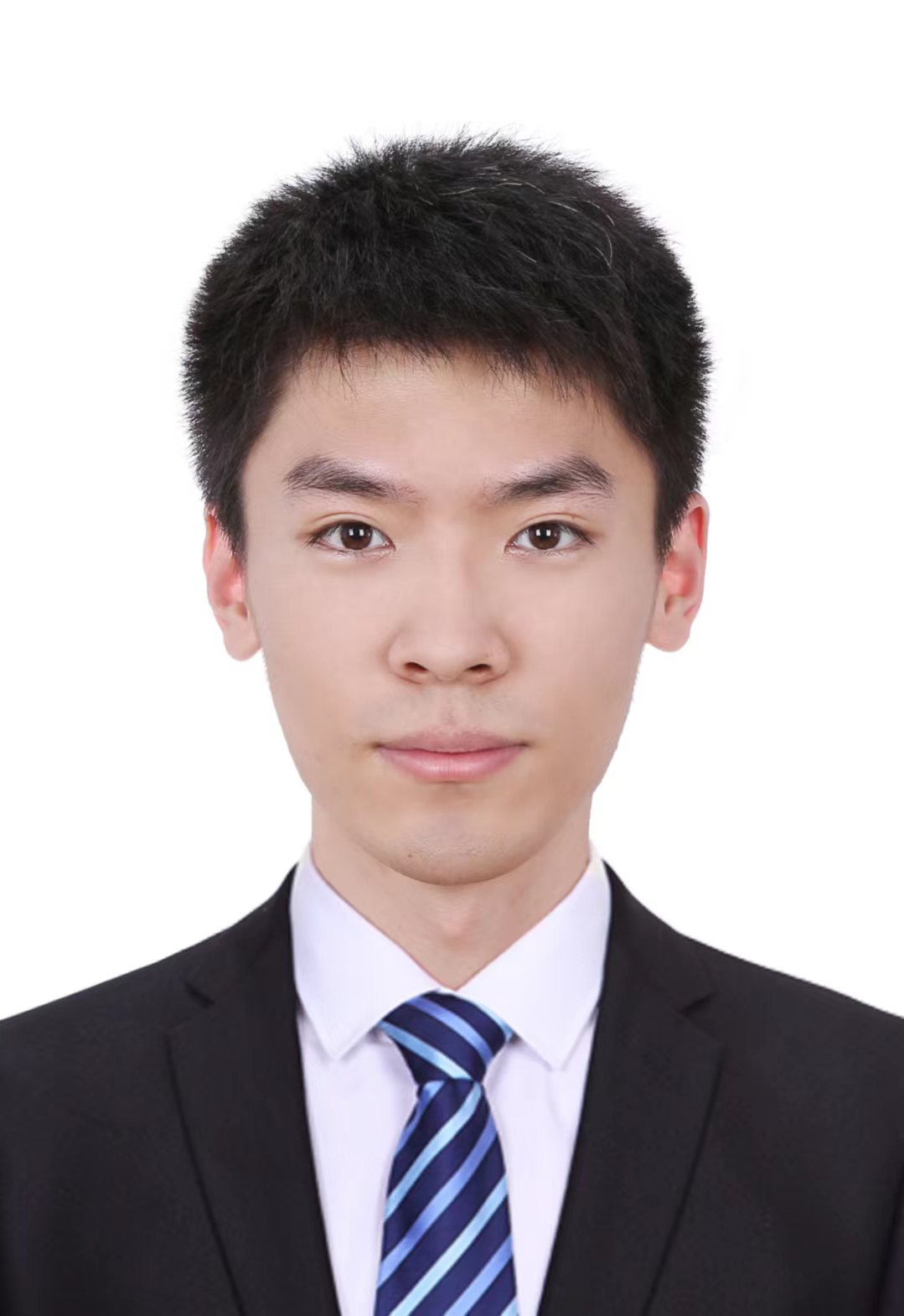}}]{Daming Shi}
 received the B.S. degree in automatic science from Beihang University, Beijing, China, in 2018 and the Ph.D. degree in automatic science from Tsinghua University, Beijing, China, in 2023. His research interests include reinforcement learning and multi-agent systems in gaming and production.
\end{IEEEbiography}
\vspace{-0.2in}
\begin{IEEEbiography}[{\includegraphics[width=1in,height=1.25in,clip,keepaspectratio]{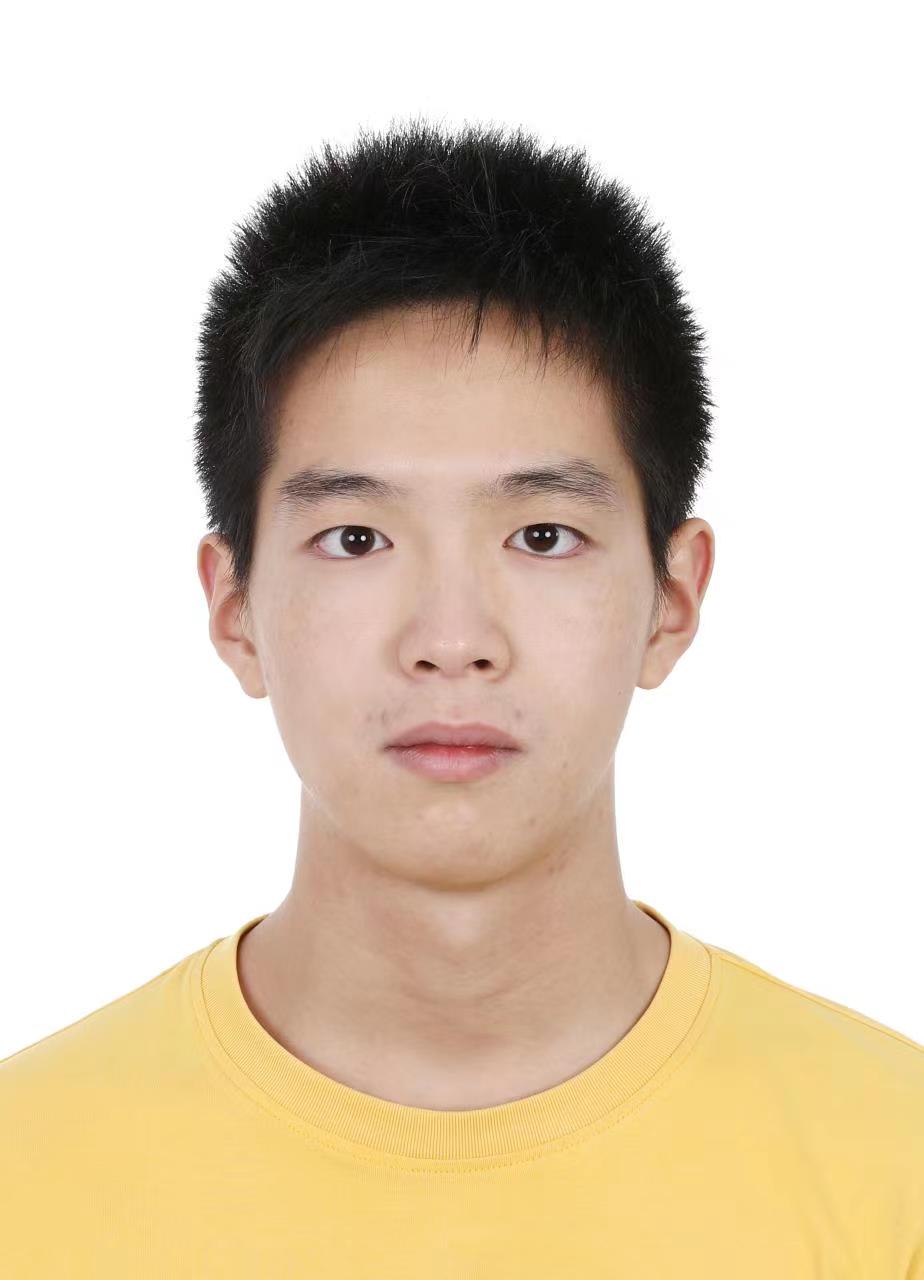}}]{Junjie Yu}
received the B.S. degree in automatic science from Beihang University, Beijing, China, in 2023. He is now pursuing the M.S. degree at the Department of Automation, Tsinghua University, Beijing, China. His current research interests include multi-agent systems and reinforcement learning. 
\end{IEEEbiography}
\vspace{-0.2in}
\begin{IEEEbiography}[{\includegraphics[width=1in,height=1.25in,clip,keepaspectratio]{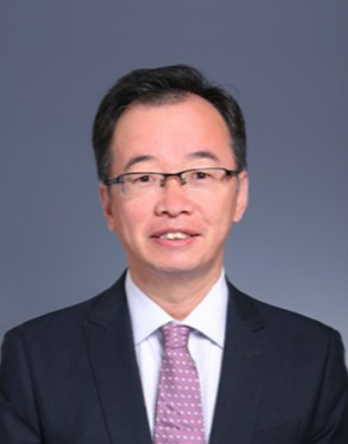}}]{Wenhui Fan}
received the Ph.D. degree in mechanical engineering from Zhejiang University, Hangzhou, China, in 1998. He obtained the postdoctoral certificate from Tsinghua University, Beijing, China, in 2000. He is a vice president of China Simulation Federation. He is currently a professor at Tsinghua University, Beijing, China. His current research interests include multi-agent modeling and simulation, large scale agent modeling and simulation, and multi-agent reinforcement learning.
\end{IEEEbiography}





\vfill

\end{document}